\documentclass[10pt,twocolumn,letterpaper]{article}

\usepackage{wacv}
\usepackage{times}
\usepackage{epsfig}
\usepackage{graphicx}
\usepackage{amsmath}
\usepackage{amssymb}
\usepackage[linesnumbered,algoruled,boxed,lined]{algorithm2e}
\usepackage{graphicx}

\usepackage{subcaption}
\usepackage{slashbox}
\usepackage{authblk}

\usepackage[normalem]{ulem}

\wacvfinalcopy 

\def\wacvPaperID{****} 

\ifwacvfinal\fi
\begin{document}

\title{Deep Representation Learning Characterized by Inter-class Separation for Image Clustering}

\author[1]{Dipanjan Das \thanks{dipanjan.da@tcs.com}}
\author[1]{Ratul Ghosh\thanks{ratulghoshr@gmail.com, Ratul Ghosh was an intern @ TCS Research}}
\author[1]{Brojeshwar Bhowmick\thanks{b.bhowmick@tcs.com}}
\affil[1]{Embedded Systems and Robotics,
TCS Research and Innovation, Kolkata, India}


\maketitle
\ifwacvfinal\thispagestyle{empty}\fi

\begin{abstract}
Despite significant advances in clustering methods in recent years, the outcome of clustering of a natural image dataset is still unsatisfactory due to two important drawbacks. Firstly, clustering of images needs a good feature representation of an image and secondly, we need a robust method which can discriminate these features for making them belonging to different clusters such that intra-class variance is less and inter-class variance is high. Often these two aspects are dealt with independently and thus the features are not sufficient enough to partition the data meaningfully. In this paper, we propose a method where we discover these features required for the separation of the images using deep autoencoder. Our method learns the image representation features automatically for the purpose of clustering and also select a coherent image and an incoherent image simultaneously for a given image so that the feature representation learning can learn better discriminative features for grouping the similar images in a cluster and at the same time separating the dissimilar images across clusters. Experiment results show that our method produces significantly better result than the state-of-the-art methods and we also show that our method is more generalized across different dataset without using any pre-trained model like other existing methods.
\end{abstract}

\begin{figure*}
\begin{subfigure}{.5\textwidth}
  \centering
  \includegraphics[width=.7\linewidth]{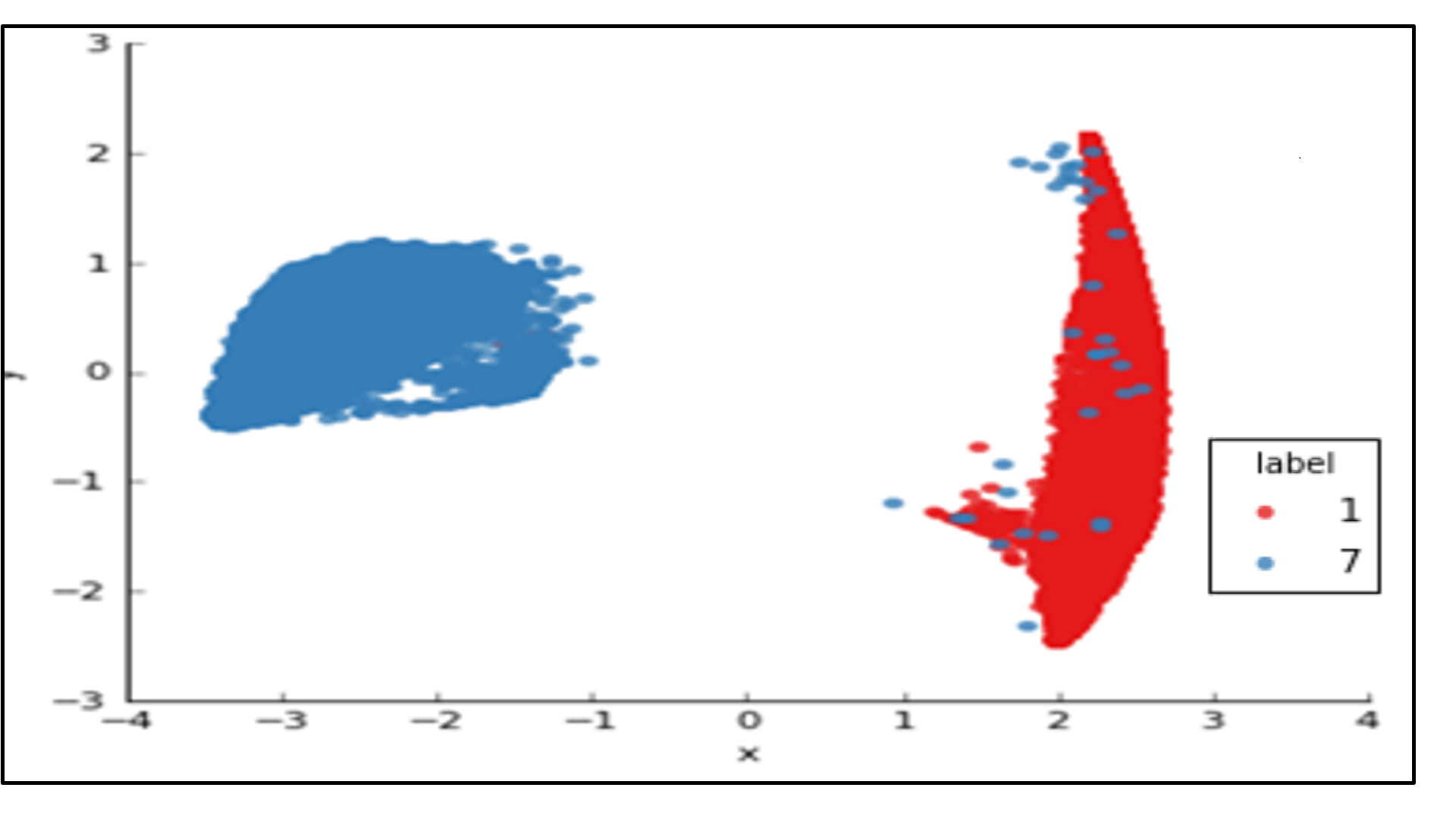}
  \caption{The distribution of latent representations for character\\ 1 and 7 from MNIST dataset using approach proposed by \\Xie \textit{et al.} \cite{xie2016unsupervised}. This demonstrates that there exists overlapping \\representation between class 1 and 7 as there are several blue \\dots present on red segment.   }
  \label{fig:Embeddings_DEC}
\end{subfigure}%
\begin{subfigure}{.5\textwidth}
  \centering
  \includegraphics[width=.7\linewidth]{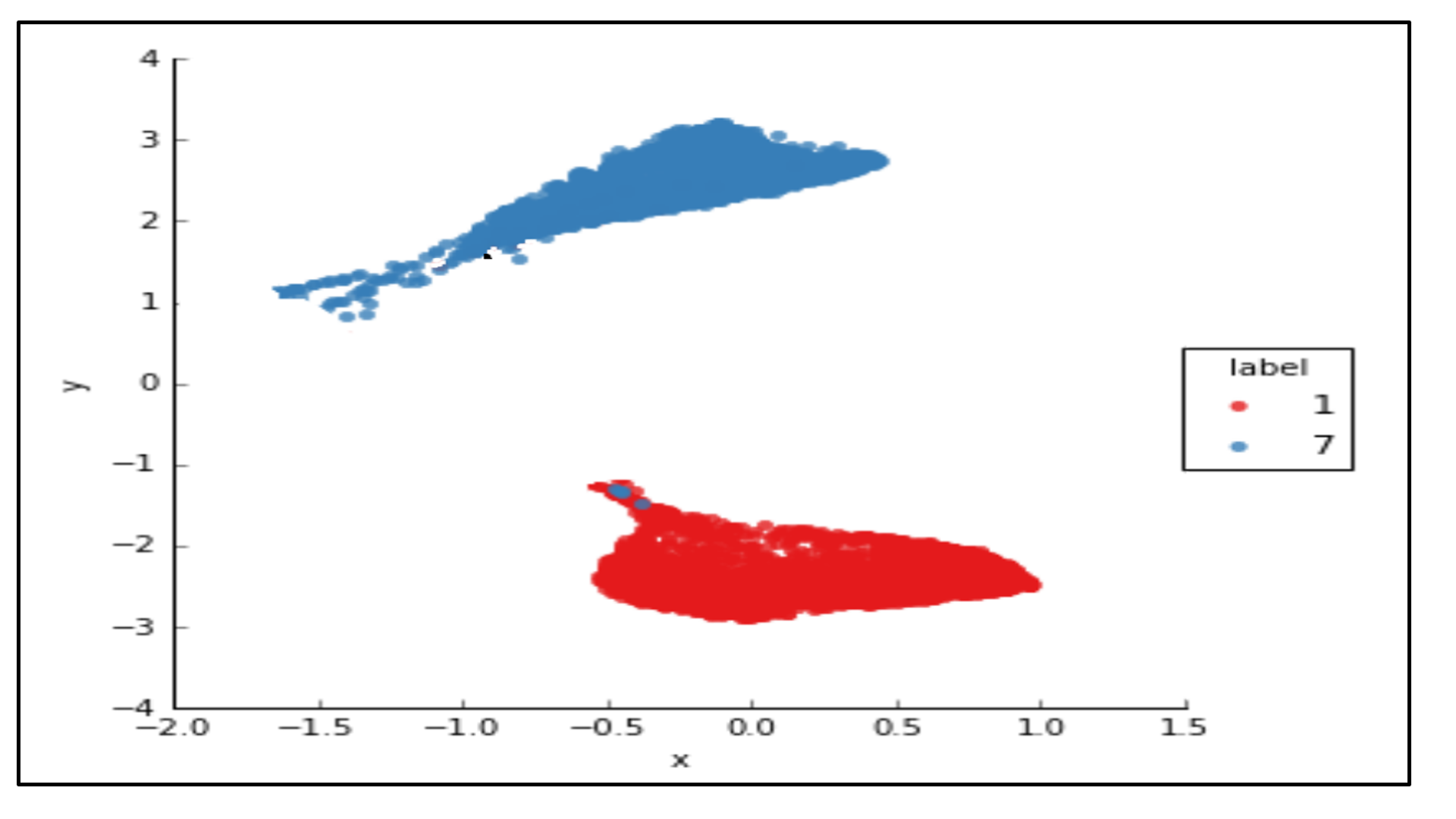}
  \caption{The distribution of latent representation for character 1 and 7 from MNIST dataset using our approach. There exists very less overlapping between class 1 and 7, only a few blue dots present on the red segment. }
  \label{fig:Embeddings_ours}
\end{subfigure}

\caption{We demonstrate the qualitative improvement of feature separation using our method. We use \textit{t-SNE} \cite{maaten2008visualizing} for visualization of 10-dimensional latent representation in 2-dimensional space (Best viewed in color).}
\label{fig:ImproveEmbeddings}
\end{figure*}
\section{Introduction}
Clustering of images is one of the fundamental and challenging problems in computer vision and machine learning. Several applications like 3D reconstruction from an image set \cite{frahm2010building}, storyline reconstruction from photo streams \cite{6909934,6909891}, web-scale fast image clustering \cite{avrithis2015web,7298596} etc. use image clustering as one of the important tools in their methodology. While the literature for clustering algorithm is voluminous \cite{shi2000normalized,ng2002spectral,ye2008discriminative,yang2010image,von2007tutorial,yang2016joint,gdalyahu2001self}, there are roughly two types of approaches for clustering viz.  hierarchical clustering and centroid-based clustering \cite{8017517}. Most popular algorithm in the hierarchical methods is agglomerative clustering which begins with many small clusters, and then merges clusters gradually \cite{gdalyahu2001self,yang2016joint} to produce a certain number of clusters. On the other hand, centroid based methods (e.g. $K$-means ) \cite{ye2008discriminative} pick $K$ samples from the input data for the initialization of the cluster centroids which are further refined iteratively by minimizing the distance between the input data and the centroids. All of these methods require the notion of feature to represent an image data so that a meaningful partition of an image dataset can be obtained. If the dimension of the feature is very high then it becomes ineffective to compute the distance between the features due to the well-known fact of curse of dimensionality \cite{pmlr-v70-yang17b}. 
Hence a variety of methods like principal component analysis (PCA), canonical correlation analysis (CCA), nonnegative matrix factorization (NMF) and sparse coding (dictionary learning) etc. are regularly used to reduce the dimensionality of features before clustering. Therefore the performance of the clustering methods crucially depends on the choice of these different feature representations.

\begin{figure*}[ht]
\centering
\includegraphics[width=\textwidth,height=5cm]{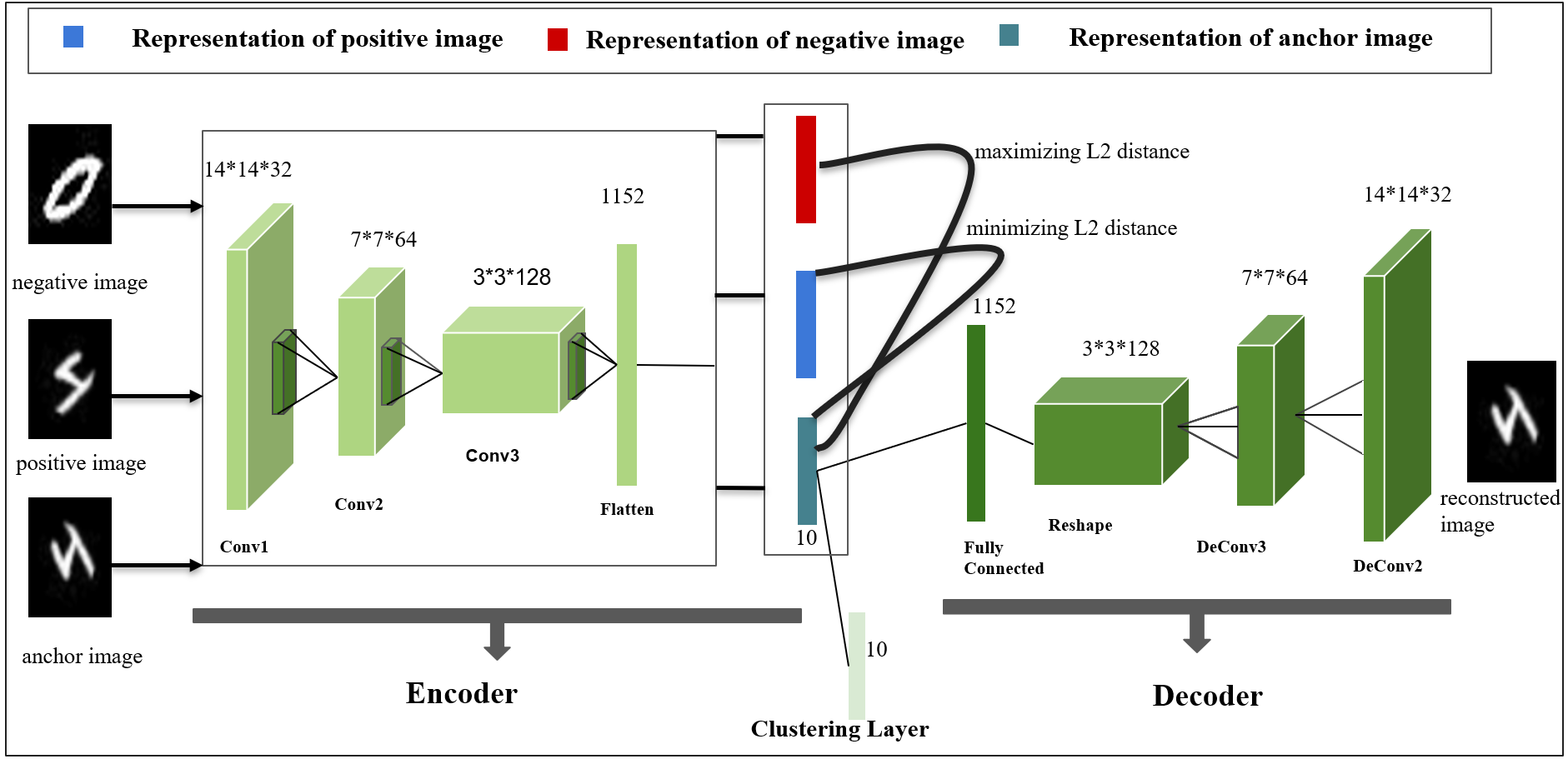}
\caption{Our deep learning based clustering network. Our network takes three images as an input during training to learn the latent representation using convolution autoencoder whose objective is to minimize the distance between an anchor and a positive image and maximize the distance between an anchor and a negative image in their latent representation space. Clustering layer uses the latent representation and assigns it to the appropriate cluster. We use decoder layer to preserve the local structure of data by minimizing the reconstruction loss between the anchor and reconstructed image (Best viewed in color).}
\label{fig:DAC}
\end{figure*}

In recent years, deep-learning (DL) methods have made an enormous success in producing good feature representation of an image which is required for different machine learning tasks \cite{he2016deep,krizhevsky2012imagenet,dahl2012context}.    
   DL learns these powerful representations from the image through high-level non-linear mapping \cite{bengio2013representation} and hence these representations can be used to partition the data into different clusters. $K$-means works better with such representations learned by DL models than using other traditional methods \cite{pmlr-v70-yang17b}. There are broadly two approaches to use the deep representation of the data for clustering. The first naive approach is to use the hidden representation (features) of the data which are extracted from a well-trained deep network \cite{tian2014learning} using supervision. However, this approach cannot fully exploit the power of deep neural network for unsupervised clustering because the usage of an already trained deep network for some other purpose lacks of knowledge of feature required for partitioning an unknown data. In the second approach an existing clustering method is embedded into a DL model. This association enables the DL model to learn cluster-oriented meaningful representation. For example, Song \textit{et al.} \cite{song2013auto} integrate $K$-means algorithm into a deep autoencoder and does cluster assignment in the bottleneck layer. Such an approach produces better outcome as the features learned by the DL models are relevant for clustering. Also, if we can predict the distribution of the cluster assignment of the features through some auxiliary distribution then, the observed cluster assignment of a feature and the auxiliary distribution together can improve the accuracy of the clustering. To this end, in a recent work, Xie \textit{et al.} \cite{xie2016unsupervised} propose a clustering objective to learn deep representation by minimizing the KL divergence between the observed cluster assignment probability and an auxiliary target distribution which is derived from current soft cluster assignment. For designing the auxiliary distribution Xie \textit{et al.} \cite{xie2016unsupervised} use a function of current model assignment distribution and the frequency per cluster. However, the cluster assignment probabilities keep on changing with training and therefore without any constraints, the auxiliary distribution also keeps on changing resulting in poor quality of representation as shown in Figure \ref{fig:Embeddings_DEC}. To improve the representation, in a recent method Hu \textit{et al.} \cite{pmlr-v70-hu17b} propose a new regularization on their DL model which enforces that feature representations of intra-class should come close to each other to improve the representation learning. They use a pre-trained model for feature enrich datasets like CIFAR10 and CIFAR100 but do not achieve comparable accuracy with other datasets like MNIST. 
  Along with the constraints like feature coherence of images for same class, clustering methods should also have an important constraint like the separation between inter-class features should also be high. But in an unsupervised setting detection of image which does not belong to same class is difficult.
  
In this work, we use a new constraint on Deep Convolutional Autoencoder (DCA) model whose objective to bring intra-class images closer and make inter-class images distant in their representation space simultaneously. To this end, for every anchor image in the database we select a positive image which is an augmented version of the anchor image and a negative image which is neither similar to the original anchor image nor the augmented image in their latent representation space. The negative image is selected from the image database using our proposed selection algorithm described in Section \ref{Representation}. We put a constraint in clustering using this image triplet (anchor image, positive image and  negative image) to minimize the distance between the anchor and the positive image and at the same time maximize the distance between the anchor and the negative image in their latent representation space. This new constraint allows the model to learn better feature representation which in turn produce a better auxiliary distribution (as this is a function of the feature representation \cite{xie2016unsupervised}) required for improved accuracy in clustering being in unsupervised setting. We also propose a method to avoid a degenerate condition where an image belonging to a class can have a considerable probability to belong to multiple clusters during training due to the unconstrained model assignment probability. 
To mitigate this, we use the difference between the entropy of the average model assignments and the entropy of model assignments inspired from ~\cite{bridle1992unsupervised} as an additional constraint, which enforces to produce unambiguous model assignments.

\begin{figure*}[t]
\includegraphics[width=\textwidth, height=6.7cm]{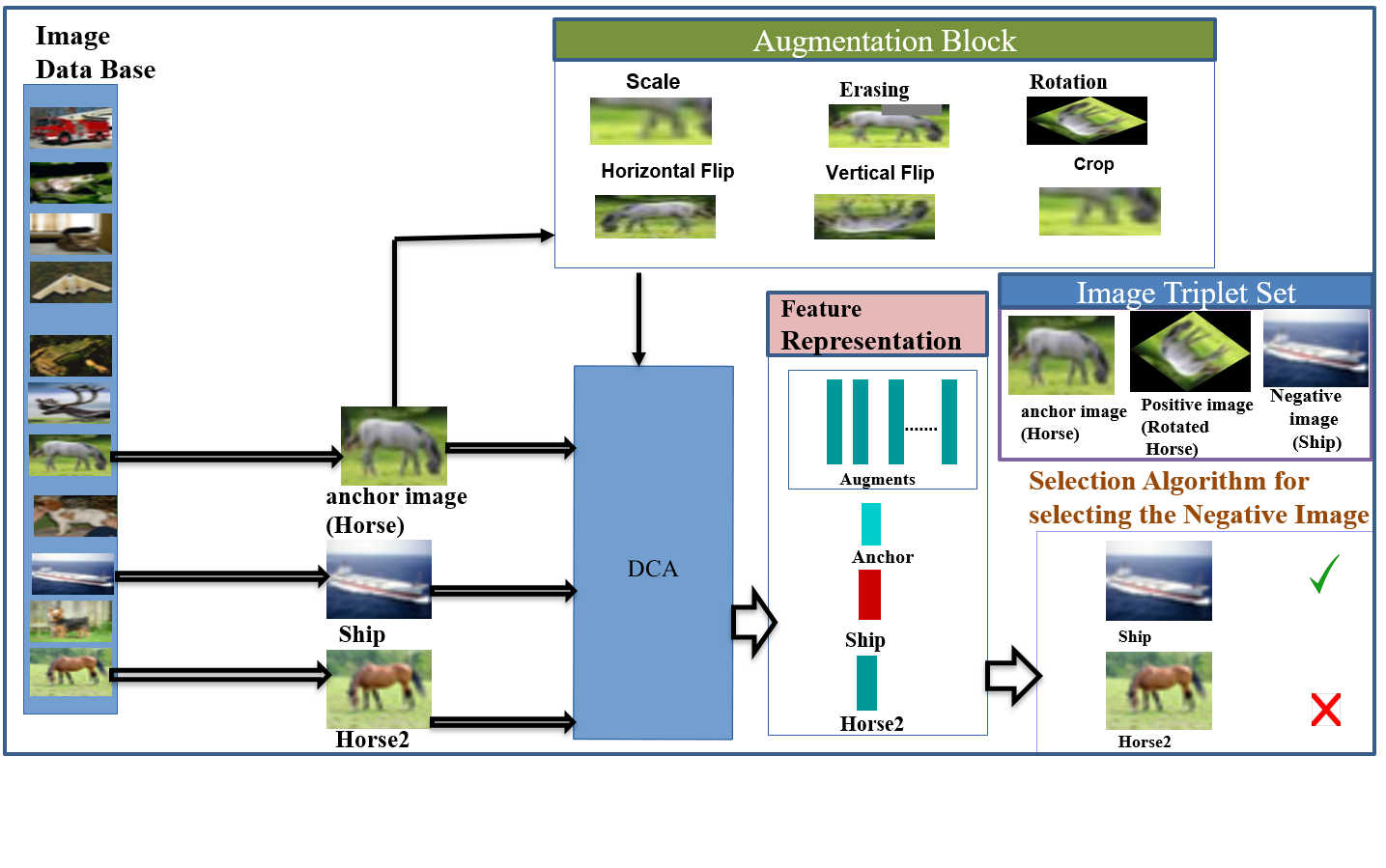}
\caption{ We demonstrate the selection process to form image set consists of anchor image, positive image and negative image.  An anchor image, for example Horse in this figure, is chosen from image database randomly. Positive image is chosen randomly from the different  augmentation of the anchor image as shown in augmentation block. Negative image is selected by our selection algorithm using the $L2$ distance between the feature representation produced by our DCA network for anchor image, its augments and negative image. For instance, our selection algorithm chooses $Ship$ image as a negative image for the anchor image horse instead of \textit{Horse2} (Best viewed in color).} 
\label{fig:Triplet}
\end{figure*}

The main contributions of this paper are :
\begin{itemize}
\item We use a new constraint on deep representation learning to enforce that feature representation can learn both intra-class similarities and inter-class separation at the same time in an unsupervised setting which helps to improve clustering.
\item Our method use entropy minimization for model assignment probability which helps the model to produce better model assignments.
\item In contrast to some other recent methods which use features from some already pre-trained models for feature rich dataset like CIFAR10, our method does not depend on such existing pre-trained models. Also our method shows excellent performance in all comparative dataset which demonstrates the generalization of our method in terms of applicability.
\end{itemize}


The rest of the paper is divided into four sections, Section \ref{Section2} present the background study on unsupervised image clustering. Section \ref{Section3} describes of our method consisting of our clustering network, inter-class separation constraint and selection algorithm. We present our experiments in Section \ref{Section4} followed by a conclusion.

\section{Related Work}\label{Section2}
Classical clustering algorithms such as $K$-means \cite{macqueen1967some}, Gaus	sian mixture models \cite{bishop2006pattern} and spectral clustering \cite{von2007tutorial}, work with original data spaces and are not very efficient in high dimensional spaces with complicated data distribution such as images \cite{pmlr-v70-yang17b}. 

Early attempts in deep learning based clustering approach first train the feature representation model and then in a separate step they use these learned features for clustering \cite{tian2014learning}. In recent years, the representation learning and clustering is done based on joint optimization on deep network \cite{xie2016unsupervised,ijcai2017-273,dizaji2017deep,pmlr-v70-hu17b,yang2016joint}. Using this joint optimization method, Yang \textit{et al.} \cite{yang2016joint} propose a new clustering model Joint Unsupervised Learning (JULE), which learns representation and clustering simultaneously. It uses agglomerative clustering and achieves a good result, but it requires high memory and computational power \cite{hsu2018cnn}. Deep Embedded Clustering (DEC) \cite{xie2016unsupervised} model simultaneously learns feature representations using an autoencoder along with cluster
assignments. The encoder is pre-trained layer-wise and then fine-tuned by a clustering algorithm using KL divergence. The problem with this approach is that their representation does not adequately address inter-class separation and hence not scalable for the high dimensional dataset.
Variational Deep Embedding (VaDE) \cite{ijcai2017-273} is a generative clustering approach using variational autoencoders. It uses variational autoencoders  and Gaussian mixture model simultaneously to model the data generative process. This method also does not generalize well as it requires a pre-trained model for high dimensional image datasets. Deep Embedded Regularized Clustering (DEPICT) \cite{dizaji2017deep} uses joint optimization to learn feature representation and cluster assignment using a convolutional autoencoder. It uses 
a regularization, which enforces that each cluster has an almost equal number of samples. But this leads to a bad clustering performance when data is not uniformly distributed across clusters, or the prior data distribution is unknown. Most recent deep learning based model is Information-Maximizing Self-Augmented Training (IMSAT)  \cite{pmlr-v70-hu17b}. It is based on Regularized Information Maximization (RIM) \cite{krause2010discriminative} which maximizes the mutual information between the input and the cluster assignment by learning a probabilistic classifier. It does augmentation to put a new constraint in its objective function where it tries to minimize the $L2$ distance between augmented data and original data in their latent representation space to understand the intra-class similarities of samples. It also requires some labeled data for augmentation and uses fixed pre-trained network layers for feature enrich dataset like CIFAR10 \cite{torralba200880} and CIFAR100 \cite{torralba200880}.

So, there are broadly two types of problems with the previous works (1) unable to learn deep representations which are characterized by both intra-class similarities and inter-class dissimilarities \cite{xie2016unsupervised,dizaji2017deep,pmlr-v70-hu17b} , (2) issue of prior knowledge of data distribution \cite{dizaji2017deep}.
In this paper, we propose new constraints for both representation learning and clustering to solve these problems.




\section{Proposed Methodology}\label{Section3}
In this section, we first introduce our convolution autoencoder and the clustering layer of our clustering network. Then, we describe in detail about the clustering constraints required for our network and their formulation. Finally, we describe the composite objective function for our network which simultaneously estimates the cluster assignment and updates the network parameters. 

\subsection{Deep Convolutional Autoencoder}
Our deep convolutional autoencoder (DCA) has an encoder layer which maps the input image $x_{i}$  to it's latent representation $z_{i} = F_\phi(x_{i})$ using a stack of convolution layers followed by a fully connected layer. The decoder layer has a fully connected layer followed by a stack of  deconvolutional layers to convert the latent representation back to original image $x_{i}' = G_\theta(z_{i}) = G_\theta(F_\phi(x_{i})) $ where $\phi$  and $\theta$ are the parameters of the encoder and the decoder respectively. The corresponding architecture is shown in Figure \ref{fig:DAC}.
We minimize reconstruction loss given by the Equation \ref{1}. The purpose of using the reconstruction loss is to preserve the local structure of the input images into features space.
 
\begin{equation}\label{1}
L_r = \frac{1}{n} \sum_{i=1}^{n} ||G_\theta(F_\phi(x_i)) - x_i||^2_2
\end{equation}
 
\subsection{Clustering Layer}
The clustering layer in our network consists of cluster centers {$\mu_j$} where $j={1....K}$, where $K$ is the number of predefined clusters. Our clustering layer is similar to the network which is proposed by \cite{xie2016unsupervised}.
Similar to \cite{maaten2008visualizing} we use \textit{student-t} distribution to measure the similarity between $z_{i}$ and the cluster center $\mu_j$. The probability (model assignment) of assigning sample $i$ to cluster $j$ is given by the Equation \ref{2}.

 
\begin{equation}\label{2}
q_{ij} = \frac{(1+|z_i-\mu_j|^2)^{-1}}{\sum_{j} (1+|z_i - \mu_j|^2)^{-1}}
\end{equation}

In absence of the target label in an unsupervised setup, we use the auxiliary probability from \cite{xie2016unsupervised} for assigning a data $i$ to a cluster $j$ as given by Equation \ref{3}.
 
\begin{equation}\label{3}
p_{ij} = \frac{q_{ij}^2/\sum_{i}q_{ij}}{\sum_{j}q_{ij}^2/\sum_{i}q_{ij}}
\end{equation}
Now, we compute the clustering loss $L_{c}$ using the  KL divergence between model assignment probability distribution and the auxiliary distribution as given by the Equation \ref{4}. 
 
\begin{equation}\label{4}
L_c =KL(P||Q)= \sum_{i}\sum_{j}p_{ij}log\frac{p_{ij}}{q_{ij}}
\end{equation}

KL divergence loss given by the Equation \ref{4}, cannot produce a better discriminative representation in an unsupervised setting where \textit{actual} target distribution is unavailable. Figure \ref{fig:Embeddings_DEC} shows that there exist an overlapping distribution of latent representations for class 1 and 7 of  MNIST. This overlapping representation occurs 
due to change in auxiliary probability distribution (Equation \ref{3}) as training progress. To mitigate this problem, we use a new constraint on representation learning to learn more discriminating feature by simultaneously minimizing the $L2$ distance between images of the same class and maximize the $L2$ distance among inter-classes. 
\subsection{ Representation Learning Using Inter-class Separation Constraint }\label{Representation}
We use a new constraint, given by the  Equation \ref{5},  whose objective is to minimize the $L2$ distance between similar images and maximize the $L2$ distance among inter-classes in their featrure representation space. This constraint enforces that the features of the anchor image ($I_a$) and the positive image ($I_p$) should come close and at the same time the distance between the features of the anchor image ($I_a$) and the negative image ($I_n$) should increase.

\begin{equation}\label{5}
L_t = \sum_{i=1}^{n}\max(0,[||z_{i}^{a} - z_{i}^{p}||^2_2 - ||z_{i}^{a} - z_{i}^{n}||^2_2 + margin])
\end{equation}
 
\begin{itemize}
 
\item $z_{i}^{a}$ is the  representation of the anchor image, $I_a$, i.e $z_{i}^{a} = F_\phi(I_a)$
\item $z_{i}^{p}$ is the  representation of the positive image, $I_p$, i.e $z_{i}^{p} = F_\phi(I_p)$
\item $z_{i}^{n}$ is the  representation of the negative image, $I_n$, i.e $z_{i}^{n} = F_\phi(I_n)$
\item $margin$ ensure the network doesn't optimize towards $z_{i}^{a}$ - $z_{i}^{p}$ = $z_{i}^{a}$ - $z_{i}^{n}$ = 0.
 
\end{itemize}
We use different augmentation methods like rotation, scale, erasing, flip, to produce the positive image ($I_p$) from the anchor image ($I_a$). It is a difficult task to select a negative image ($I_n$) for the anchor image ($I_a$) in an unsupervised setting where labels are unknown. Random selection of a negative image ($I_n$) from the image dataset cannot guarantee that it is truly negative with respect to the anchor image ($I_a$) and therefore may fall into the same class that of the anchor image ($I_a$).
So, we propose an algorithm to select the proper negative image ($I_n$) to enforce that negative image ($I_n$) does not belong to the same class of the anchor image ($I_a$) to satisfy the necessary condition for the Equation \ref{5}. 


Our selection algorithm tries to discover the difference between the distribution of the augmented images and the anchor image in their latent representation space by using $L2$ distance. It selects the appropriate image as a negative image from the dataset whose latent representation does not fall within the distribution of the augmented images of the anchor image. Algorithm (\ref{Algo}) describes the complete algorithm for selecting a negative image. We pre-trained our DCA for three epochs using reconstruction loss to initialize the parameter $\phi$ of the encoder to enable our Algorithm (\ref{Algo}) to use the latent representation for selecting negative images.

\begin{algorithm}[h]
\caption{Selection algorithm}\label{Algo}
\SetAlgoLined
\KwIn{ anchor image $I_a$, augments of $I_a$ [$I_p^{i}$ where $i=1$ to $10$,]   image dataset $I$, $\phi$  }
/* 10 augments based random rotation, scale, erasing, flip, brightness and contrast etc */\; 
$(I_p^{i})$ = augment($I_a^i$) where $i=1$ to $10$  \; 
$(Z_p^{i})$ = $F_\phi$($I_p^{i}$) where $i=1$ to $10$ \;
$z_a$ = $F_\phi$($I_a$)\;
 \While{negative image($I_n$) not found}{
   $I_n$ = select random image from $(I)$  \\
   $z_n$ = $F_\phi$($I_n)$ \\
  \eIf{$||z_a - z_n|| > $ $max $ $(||z_a - Z_p^{i}||)$}{
   return $I_n$\;
   break\;
   }{
   continue\;
  }
 }
 
\end{algorithm}
 
\begin{figure*}
\begin{subfigure}{.5\textwidth}
  \centering
  \includegraphics[width=.5\linewidth]{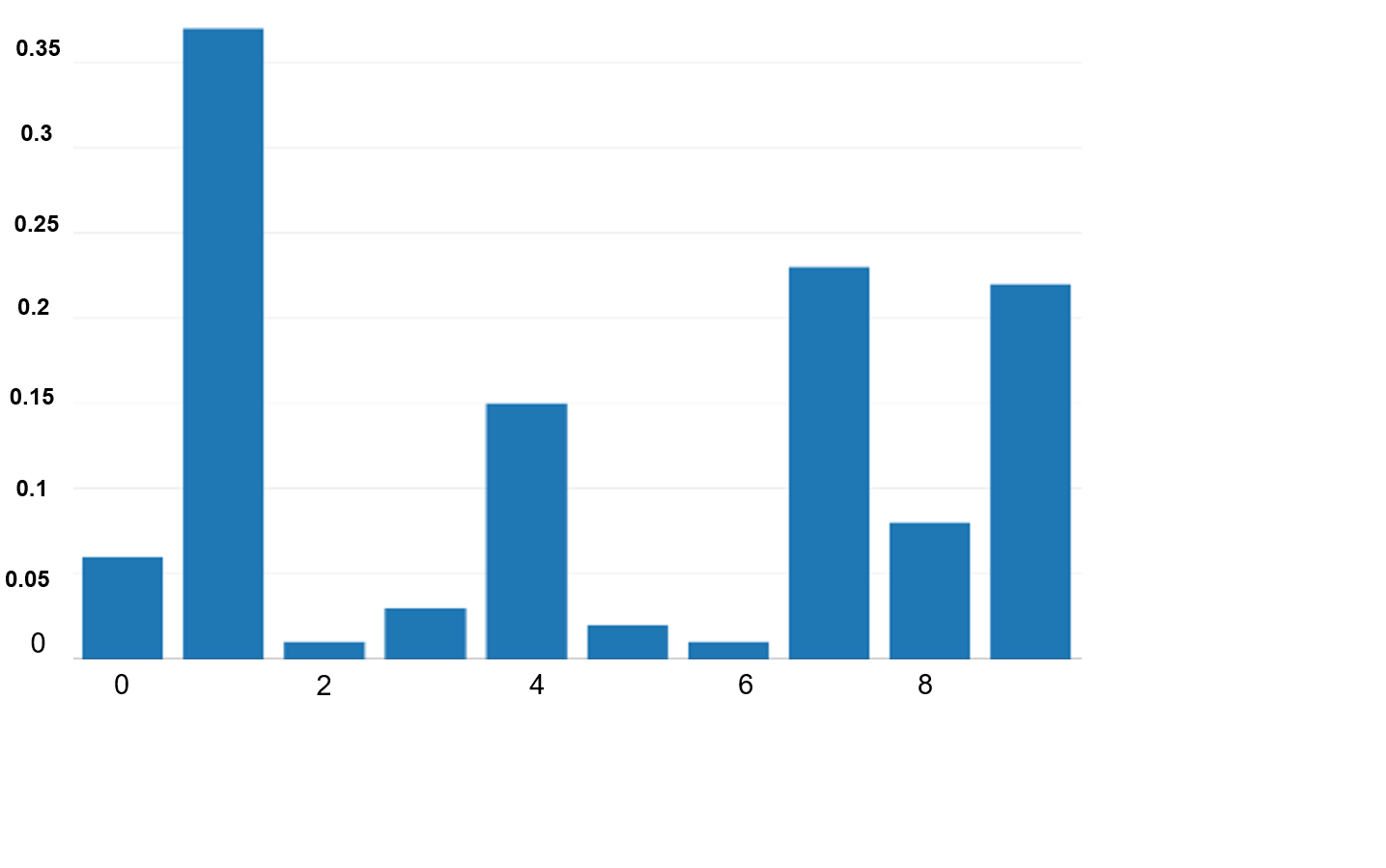}
  \caption{}
  \label{fig:PBD_Bad_for_1cls}
\end{subfigure}%
\begin{subfigure}{.5\textwidth}
  \centering
  \includegraphics[width=.5\linewidth]{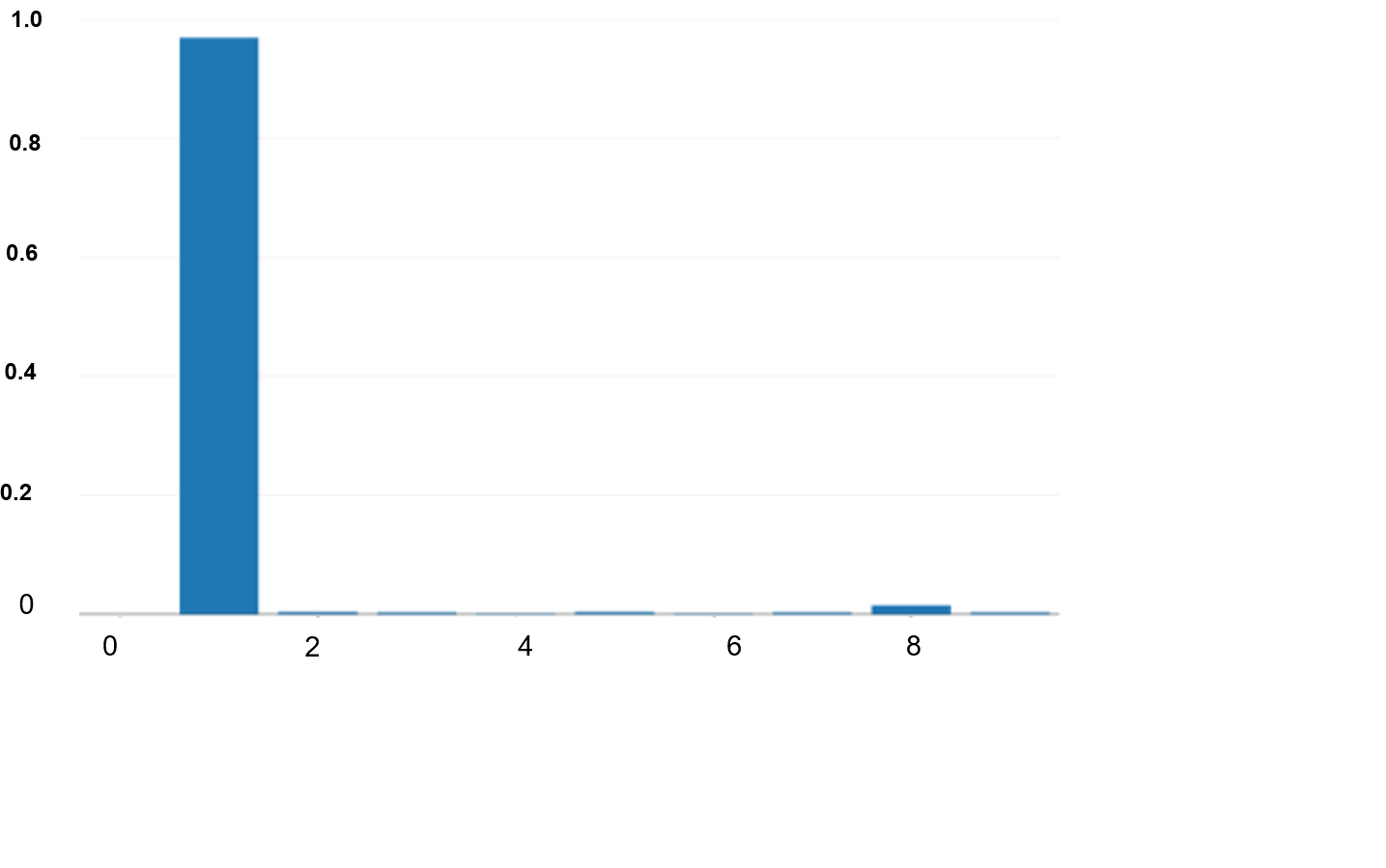}
  \caption{}
  \label{fig:PBD_Good_for_1cls}
\end{subfigure}
\caption{We demonstrate the impact of the constraint $L_b$ on the model assignment distribution of class 1 of MNIST . Here X-axis represents the class and Y-axis represents the probability. a) shows the ambiguous cluster assignments of images of class 1 because the probability of assigning that to class 9, 7 and 4 are also high when we do not use $L_b$. b) shows that images of class  $1$ have the only probability to belong to a single cluster only which is their own class after using $L_b$.}
\label{fig:fig:PBD_Bad_for_1cls_ALL}
\end{figure*}

Figure \ref{fig:Triplet} demonstrates our selection process. 
Such a method enables the network to learn better representation which helps in improvement clustering accuracy. However, an image belonging to a class can have a considerable probability to belong to multiple clusters during training due to the unconstrained model assignment probability over classes as mentioned in Equation~\ref{2}. For instance, Figure \ref{fig:PBD_Bad_for_1cls} shows that for MNIST dataset, $35$\% of the images from class $1$ have high probabilities for belonging to its own class and rest of the samples from class $1$ belongs to other classes during training after $10$ epoch. This leads to an incorrect clustering without explicit constraint as shown in our ablation study in Figure \ref{fig:MNIST}. To mitigate this, we use a constraint, given by the Equation \ref{6}, based on the difference between the entropy of the average of the $q_{ij}$ (Equation \ref{2}), and the entropy of the $q_{ij}$ (Equation \ref{2}) so that we can retain the information of the input in $q_{ij}$  similar to ~\cite{bridle1992unsupervised}. Figure \ref{fig:PBD_Good_for_1cls} shows the histogram of the images from class $1$ from MNIST after adding this constraint (Equation \ref{6}) which clearly indicate that the class assignment becomes significantly better during training resulting in improved accuracy overall. The corresponding ablation study is shown in Figure \ref{fig:MNIST}.

%

\begin{equation}\label{6}
\begin{split}
L_b = (\sum_{i}\sum\limits_{j=1}^K q_{ij} log(q_{ij})) - (\sum\limits_{j=1}^K \bar{q_{ij}} log(\bar{q_{ij}})) \\
\text{where  } \bar{q_{ij}} = \frac{1}{B} \sum_{i}^{B}q_{ij}
\end{split}
\end{equation}
Here $B$ is the batch size and $q_{ij}$ is taken from the Equation ~\ref{2}.


 
 
 
\subsection{Objective Function}
The final loss function for our network given by the Equation \ref{8}.
\begin{equation}\label{8}
L = \alpha L_r + \beta L_c + \gamma L_b + \omega L_t
\end{equation}
with the hyper parameter of $\alpha$ , $\beta$ , $\gamma$ ,  $\omega$. We use all these constraints  to optimize our network. We show in our ablation study the efficacy of each of the terms.
 
\section{Experiments}\label{Section4}
 
\subsection{Datasets}
We evaluate our proposed clustering network on the following, widely-used image datasets:
\begin{itemize}
 
\item The MNIST dataset \cite{lecun1998gradient} consists of total 70000 handwritten digits of $28\times28$ pixels. We use all images from the training and test sets without their labels.
\item USPS: It is a handwritten digits dataset from the USPS postal service \cite{291440}, containing 11,000 samples of $16\times16$ images.
 
\item CIFAR-10  \cite{torralba200880} contains $32\times32$ color images of ten different object classes. Here, we use all images of the training set.
 
\item CIFAR-100 \cite{torralba200880} contains $32\times32$ color images of hundred different object classes. Here, we use all images of the training set.
 
\item FRGC  \cite{yang2016joint}: We use the method proposed by \cite{yang2016joint} to select 20 random  subjects from the original dataset and collect 2,462 face images. We crop the face regions and resize them into $32\times32$ images.

\item ImageNet \cite{deng2012ilsvrc}: We select 10 random classes from the original dataset and collect 13000 images. We crop and resize them into $32\times32$ images.
 
\end{itemize}
 \begin{figure*}[h]
\begin{subfigure}{.5\textwidth}
  \centering
  \includegraphics[width=\linewidth]{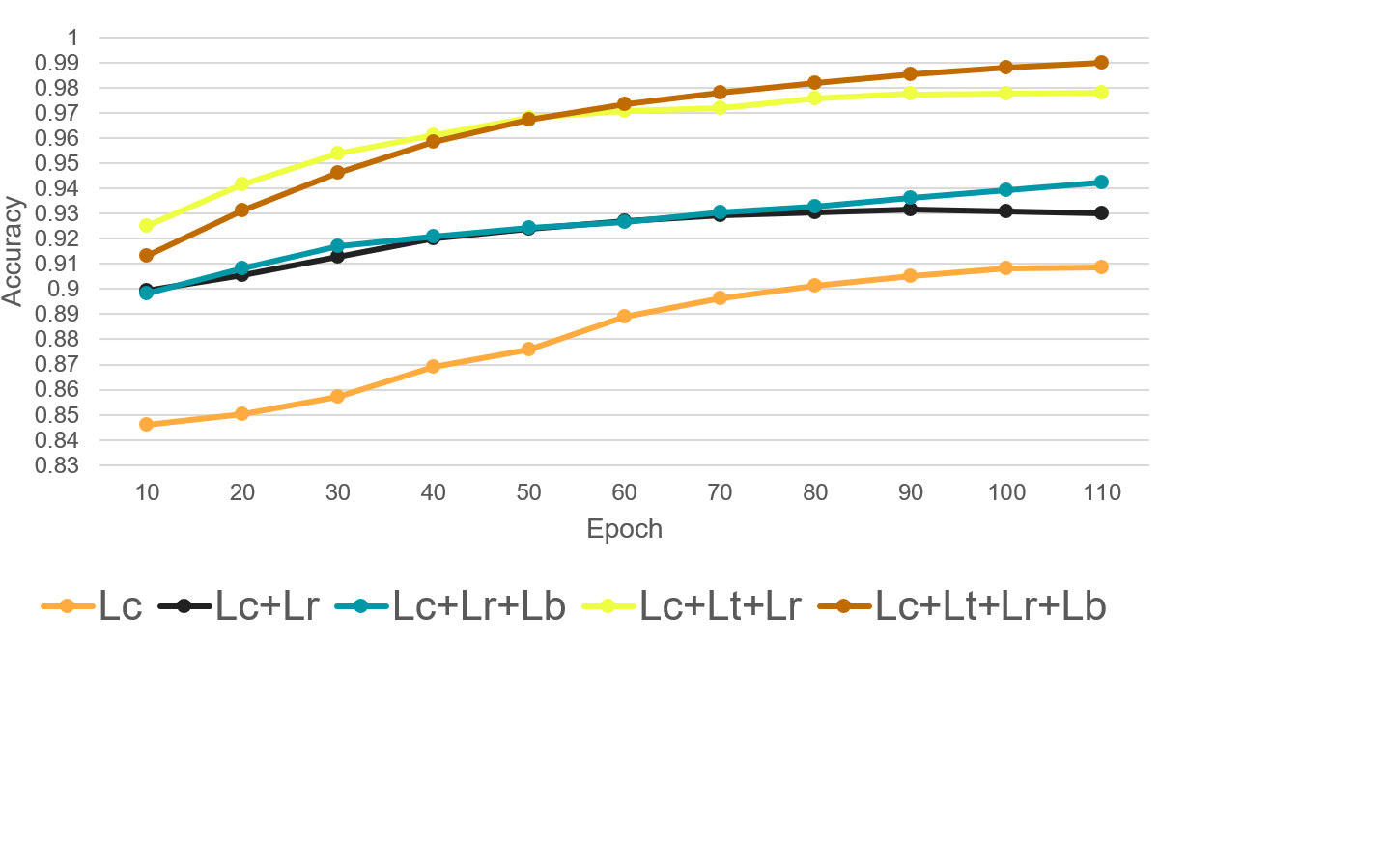}
  \caption{Result of ablation study on MNIST}
  \label{fig:MNIST}
\end{subfigure}%
\begin{subfigure}{.5\textwidth}
  \centering
  \includegraphics[width=\linewidth]{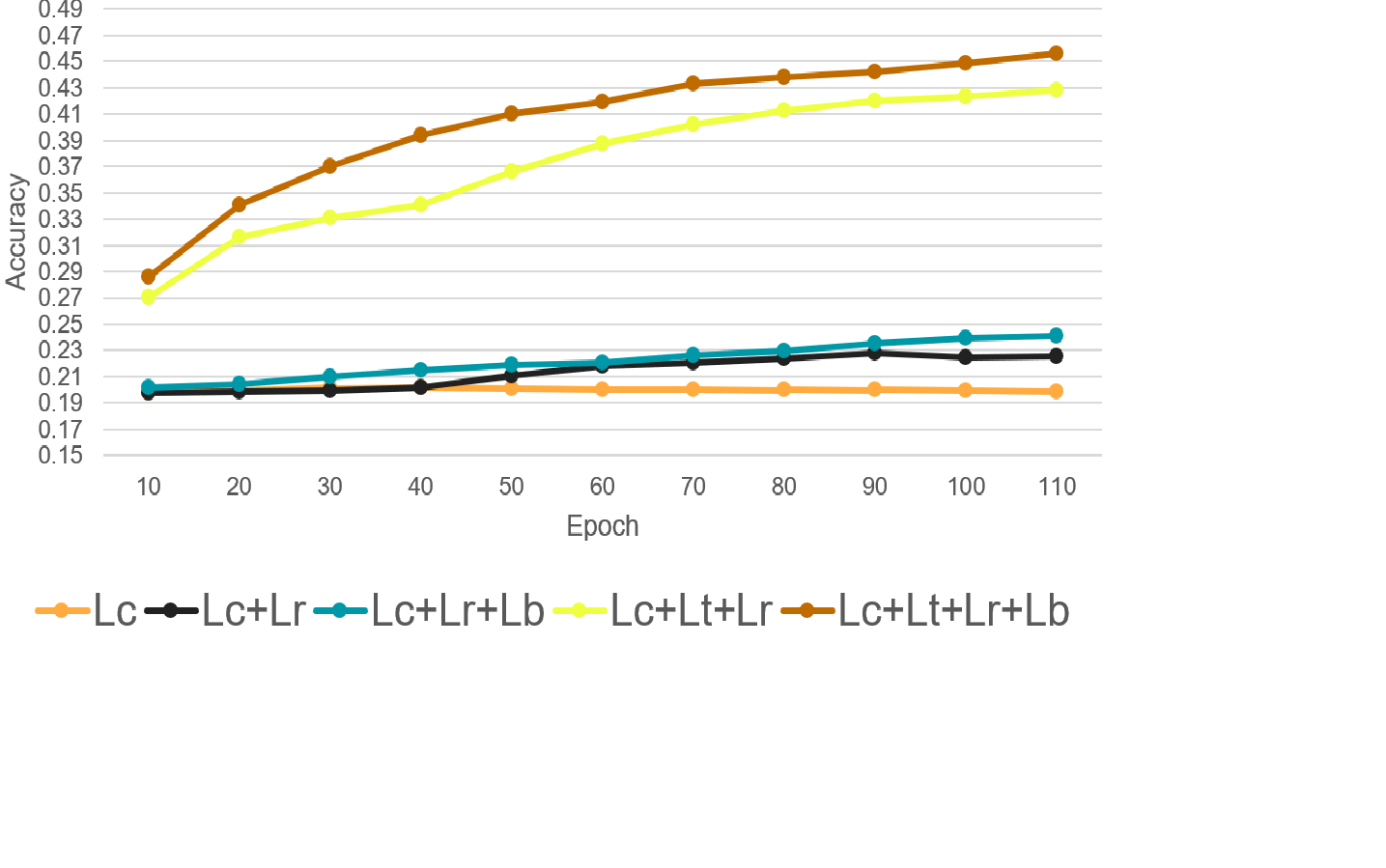}
  \caption{Result of ablation study on CIFAR10.}
  \label{fig:CIFAR}
\end{subfigure}
\caption{We demonstrate the result of our ablation study on MNIST and CIFAR10. We trained and evaluated a series of models, each having one or more constraints of the loss function mentioned in Equation \ref{8}.}
\label{fig:Ablation}
\end{figure*}

 \begin{table*}[h!]
\begin{center}
\tabcolsep 18.8pt
    \small
 \begin{tabular}{|c|| c| c| c| c| c|c|} 
 \hline
  & MNIST & USPS & CIFAR10 & CIFAR100 & FRGC & ImageNet  \\
 \hline\hline
 $K$-means on pixels & 53.49 & 46 & 20.4 & 16.5 & 24.3 & 12.65  \\ 
 \hline
 DEC \cite{xie2016unsupervised} & 84 & 62* & - & - & 37.8 &- \\
 \hline
 VaDE \cite{ijcai2017-273} & 94.46 & - & - & - &-&- \\
 \hline
 JULE \cite{yang2016joint} & 96.4 & 95* & - & - & 46.1 &- \\
 \hline
  DEPICT \cite{dizaji2017deep} & 96.5 & 96.4 & - & - & 47 &-\\
 \hline
 IMSAT \cite{pmlr-v70-hu17b} & 98.4 & - & 45.6$\dagger$ & 27.5$\dagger$ & - &- \\
 \hline
 Ours & \textbf{98.93} & \textbf{97.63} & 44.19 & 25.4 & \textbf{47.28} & \textbf{35.69} \\
 \hline

\end{tabular}
\caption{Clustering performance based on accuracy (\%), given by the Equation \ref{9}, of different algorithms(higher is better). We report the mean of 20 runs. \newline $\dagger$ denotes using weights of pre-trained deep residual networks\cite{he2016deep}. Results marked with * are excerpted from [DEPICT]. We put dash marks (-) where method does not provide any result for the dataset.}
\end{center}
\end{table*}

 \begin{table*}
\begin{center}
 \begin{tabular}{|l| c| c| c| c| c|} 
 \hline
   \backslashbox{methods} {$r_{min}$}  & 0.1 & 0.3 & 0.5 & 0.7 & 0.9 \\ [1.3ex] 
 \hline
$K$-means on pixels & 46.96 & 48.73 & 52.86 & 53.16 & 53.39  \\ 
 \hline
 \hline
DEC  & 70.10 & 80.92 & 82.68 & 84.69 & 85.41  \\ 
 \hline
 \hline
Ours  & \textbf{88.02} &\textbf{94.5} & \textbf{96.12} & \textbf{97.2 }& \textbf{97.91}  \\ 
 \hline

\end{tabular}

\caption{Clustering accuracy (\%), given by the Equation \ref{9}, on an imbalance sub-sample of MNIST.}
\end{center}
\end{table*}
\subsection{Evaluation Metric}
The clustering method is evaluated by the unsupervised clustering accuracy(ACC)  \cite{xie2016unsupervised}.
 	
\begin{equation}\label{9}
ACC = max_m\frac{\sum_{i=1}^{n}1\{l_i == m(c_i)\}}{n}
\end{equation}
where $l_i$ is is the ground-truth label, $c_i$ is the cluster assignment i.e $c_i$ = $argmax_j$($q_{ij}$) and $m \in M$ are all possible one-to-one mappings between clusters and labels.


\begin{figure*}[h]
\begin{subfigure}{.35\textwidth}
  \centering
  \includegraphics[width=.8\linewidth]{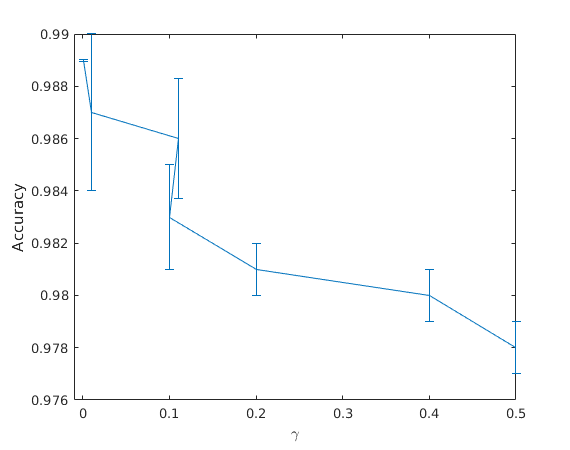}
  \caption{Selection of Hyper parameter $\gamma$.}
  \label{fig:gamma}
\end{subfigure}
\begin{subfigure}{.35\textwidth}
  \centering
  \includegraphics[width=.8\linewidth]{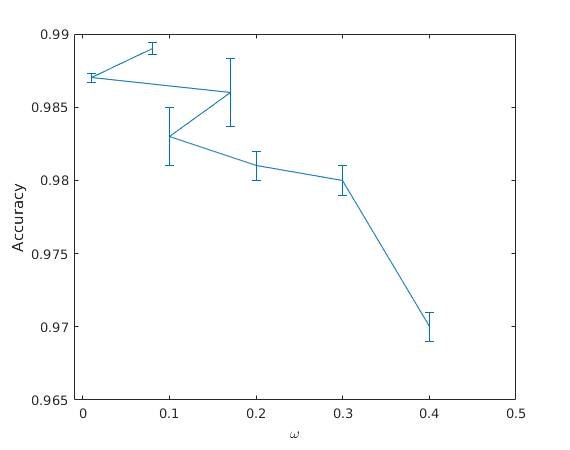}
  \caption{ Selection of Hyper parameter $\omega$.}
  \label{fig:omega}
\end{subfigure}
\begin{subfigure}{.35\textwidth}
  \centering
  \includegraphics[width=.8\linewidth]{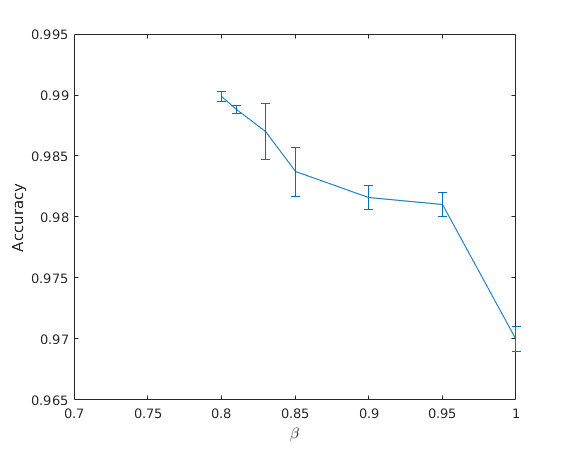}
  \caption{Selection of Hyper parameter $\beta$.}
  \label{fig:beta}
\end{subfigure}
\caption{Hyper parameters selection study. The X-axis represents the hyper parameter value, and Y-axis represents the accuracy of our network using Equation \ref{9}. The vertical line shows the range of accuracy achieved by our network where one hyper parameter (X-axis) is unchanged and the rest of the two hyper parameters are varied.}
\label{fig:HP}
\end{figure*}
 
 
\subsection{Experiment Result}

We use Tensorflow \cite{tensorflow2015-whitepaper} to implement our method on a Linux based workstation equipped with GPU. We report our result based on two types of networks, viz a shallow network suitable for the datasets like MNIST, USPS and a deeper variant of the shallow network with more parameters for the datasets like CIFAR10, CIFAR100, FRGC and ImageNet which have more features than MNIST or USPS. The encoder of the shallow network consists of three convolutional layers, with kernel size $3\times3$ and stride $1$, followed by a fully connected layer. All the convolutional layers associate with a max-pooling layer with stride $2$. The decoder of the shallow network consists of a fully connected layer followed by three deconvolutional layers. The deeper variant of our shallow network uses additional two convolution and two deconvolution layers for the encoder and the decoder respectively. For both the networks we apply a zero padding for all the convolutional and max-pooling layers. We use rectified linear units \cite{nair2010rectified} for all the hidden activations and apply batch normalization \cite{pmlr-v37-ioffe15} to each layer.  
For initialization of weights of all the layers we choose the Xavier approach \cite{glorot2010understanding}. We select Adam \cite{Kingma2014AdamAM} optimizer for the back propagation with learning-rate of $0.001$. The feature dimensions of our shallow and deep network are $64$ and $10$ respectively. We normalize the input mages to zero-mean and unit variance and pre-trained the network with only reconstruction loss to initialize the cluster centers using $K$-means.


Table 1 shows the complete results for all the datasets using our method. 
Our method produces significantly better accuracy on MNIST, USPS, and FRGC. We have achieved the competitive result on CIFAR10 and CIFAR100. IMSAT gives the best accuracy on CIFAR10 and CIFAR100 by using the features of an already pre-trained model of deep residual networks \cite{he2016deep}. This deep residual network \cite{he2016deep} is trained on ImageNet dataset of natural images which is very similar to CIFAR (dataset of natural images). Features extracts for CIFAR10 and CIFAR100 from this model have good representation which is very helpful for clustering but they do not show any result on ImageNet itself. We use the same pre-trained model on FRGC for the clustering and have obtained very poor accuracy (24.31\%) which shows that use of a pre-trained model in a method do not generalize well. In contrast, we use our network for FRGC, CIFAR10, CIFAR100 and ImageNet. We exceed state-of-the-art accuracy on FRGC and comparable accuracy with respect to IMSAT on CIFAR. To the best of our knowledge we are the first to try out ImageNet dataset for clustering. We show that in general, our method produces significantly better results with more generalization across datasets.  However, there is a substantial difference between results on the datasets like MNIST, USPS and the feature rich datasets (CIFAR10, CIFAR100, FRGC, ImageNet). This difference is due to the fact that the auxiliary probability distribution is not very close to the true distribution of data which impacts the accuracy of the feature rich dataset. 

\subsection{Ablation Study}
To understand the importance of each constraint in our method, we train and evaluate a series of models, each eliminating one or more constraints from Equation \ref{8}. We choose MNIST and CIFAR10 datasets for the ablation study. The results of our ablation study shown in Figure \ref{fig:Ablation}. It is evident from this study that the inter-class separation constraint ($L_{t}$) has a significant influence in achieving the improved accuracy.
We also conduct an extensive experiment on MNIST to find the suitable values for the hyper parameters used in Equation \ref{8}. We sample the hyper parameters in a range of $[0;1]$ and experiment on the shallow network for $100$ times. We compute the accuracy based on the evaluation metric given by the Equation \ref{9}. We keep one hyper parameter $\alpha$ fixed as $1$ always and change the rest of the hyper parameters as shown in Figure \ref{fig:HP}. Based on this experiment we choose hyper parameters for MNIST and USPS as $\alpha = 1$ , $\beta = 0.8$ , $\gamma = 0.01$ , $\omega = 0.1$. We conduct the same experiment for deeper version of the network for CIFAR10, CIFAR100, FRGC, ImageNet and find the hyper parameters as $\alpha = 1$ , $\beta = 0.7$ , $\gamma = 0.01$ , $\omega = 0.3$. We set $margin$, used in Equation \ref{5}, as 0.24 and 0.3, for shallow and deeper network respectively.

\subsection{Effect of Imbalanced Data}
In order to check the performance in an imbalance dataset, we use a similar approach proposed by \cite{xie2016unsupervised} for MNIST dataset. We use retention rate $r$ to keep the sample for a given class of MNIST. For example, if minimum retention rate is $r_{min}$, then it implies that we randomly select samples from class 0 with the probability of $r_{min}$ and class 9 with the probability of 1, and samples from other classes fall linearly in between  $r_{min}$ and 1. All the recent methods, mentioned in the Table 1, do not provide any result on an imbalance dataset except DEC. Table 2 shows that our method produces significantly improved accuracy than DEC \cite{xie2016unsupervised} for an imbalance dataset as well.

\section{Conclusion}
In this work, we propose a deep learning based clustering method which learns the high intra-class similarity and low inter-class similarity simultaneously in the latent representation space. We carefully design our selection algorithm to choose a negative image for an anchor image in an unsupervised setting. Owing to our careful choices of the constraints, the model assignment probability in our method is improved which in turn improve the auxiliary distribution resulting in improved accuracy in clustering. Improving the auxiliary distribution further through a separate learning strategy can be an immediate future work of our method for
even better clustering for complex dataset like ImageNet.

{\small
\bibliographystyle{ieee}
\bibliography{egbib}
}

\end{document}